\documentclass[letterpaper]{article} 
\usepackage{aaai2026}
\usepackage{multirow} 
\usepackage{makecell}
\usepackage{latexsym}
\usepackage{booktabs}
\usepackage{amssymb}
\usepackage{amsthm}
\usepackage{amsmath}
\usepackage{amsfonts}
\usepackage{bm}

\usepackage{times}  
\usepackage{helvet}  
\usepackage{courier}  
\usepackage[hyphens]{url}  
\usepackage{graphicx} 
\usepackage{natbib}  
\usepackage{caption} 
\usepackage{algorithm}
\usepackage{algorithmic}

\usepackage{newfloat}
\usepackage{listings}
\newtheorem{theorem}{Theorem}

\newcommand{\cmp}{x_{*}} 
\newcommand{\E}{\mathbb{E}}

\DeclareCaptionStyle{ruled}{labelfont=normalfont,labelsep=colon,strut=off} 
\floatstyle{ruled}
\newfloat{listing}{tb}{lst}{}
\floatname{listing}{Listing}
\title{Scaling and Transferability of Annealing Strategies \\ in Large Language Model Training}
\author{
    Siqi Wang\textsuperscript{\rm 1},
    Zhengyu Chen\textsuperscript{\rm 2},
    Teng Xiao\textsuperscript{\rm 3},
    Zheqi Lv\textsuperscript{\rm 2},
    Jinluan Yang\textsuperscript{\rm 2},
    Xunliang Cai\textsuperscript{\rm 2}, \\
    Jingang Wang\textsuperscript{\rm 2},
    Xiaomeng Li\textsuperscript{\rm 1, 4}
}

\affiliations{
    \textsuperscript{\rm 1}The Hong Kong University of Science and Technology, Hong Kong SAR \\
    \textsuperscript{\rm 2}Meituan Inc. \\
    \textsuperscript{\rm 3}Allen Institute for Artificial Intelligence, USA\\
    \textsuperscript{\rm 4}Shenzhen Loop Area Institute \\[2mm]
    swanggz@connect.ust.hk, chenzhengyu04@meituan.com, eexmli@ust.hk \\
}

\begin{document}

\maketitle

\begin{abstract}
Learning rate scheduling is crucial for training large language models, yet understanding the optimal annealing strategies across different model configurations remains challenging. In this work, we investigate the transferability of annealing dynamics in large language model training and refine a generalized predictive framework for optimizing annealing strategies under the Warmup-Steady-Decay (WSD) scheduler. Our improved framework incorporates training steps, maximum learning rate, and annealing behavior, enabling more efficient optimization of learning rate schedules. Our work provides a practical guidance for selecting optimal annealing strategies without exhaustive hyperparameter searches, demonstrating that smaller models can serve as reliable proxies for optimizing the training dynamics of larger models. We validate our findings on extensive experiments using both Dense and Mixture-of-Experts (MoE) models, demonstrating that optimal annealing ratios follow consistent patterns and can be transferred across different training configurations. 
\end{abstract}

\begin{links}
    \link{Code}{https://github.com/xmed-lab/fm-annealing}
\end{links}

\section{Introduction}

Previous work \citep{Kaplan2020ScalingLF, Hoffmann2022TrainingCL, rae2021scaling} on scaling laws has focused mainly on the general relationship of model performance and model size, compute budget or tokens. However, recent studies \citep{tissue2024scaling, hagele2024scaling} have been focused on the discrepancies introduced by training dynamics with different hyperparameter settings.

\begin{figure*}[t]
\centering
\includegraphics[width=0.7\textwidth]
{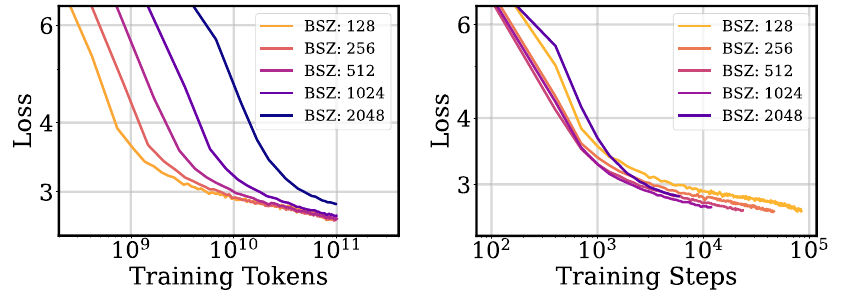}
\caption{\emph{(Left)} Loss vs Training tokens across different batch sizes with sequence length as 8192. \emph{(Right)} Loss vs Training steps across different batch sizes with sequence length as 8192. The standard deviation of the loss values for batch sizes 256, 512, and 1024 is approximately 0.00847.}
\label{fig:loss_tokens_bsz}
\end{figure*}

When training a model, even if the amount of total tokens remains constant, different settings can lead to different loss trajectory over time (as shown in Figures~\ref{fig:loss_tokens_bsz}). This suggests that the path the model follows to reach its irreducible loss varies depending on multiple hyperparameters in the training setup. Apart from batch sizes and maximum learning rates, a learning rate scheduler with warmup and annealing stages is crucial for achieving optimal performance and generalization in language models \cite{He2015DeepRL, Goyal2017AccurateLM, Popel2018TrainingTF, You2019LearningSP}.

Firstly, our work aims to address the first question: beyond training tokens and model sizes, what factors influence the dynamics of the loss curve across different training settings? In our work, we find that training steps and annealing strategies play a crucial role, and we also observe that different batch sizes lead to distinct training loss curves (shown in Figure~\ref{fig:loss_tokens_bsz} \emph{(Left)}). While large batch sizes are often recommended for various tasks \cite{You2017100epochIT, You2017LargeBT, Smith2017DontDT, Puri2018LargeSL, Kaplan2020ScalingLF} to improve efficiency, we observe in Figure~\ref{fig:loss_tokens_bsz} that once the batch size exceeds a certain threshold, the training loss curves with respect to training steps tend to converge. \citet{mccandlish2018empirical} proposed the concept of an optimal batch size, which is the smallest size that effectively approximates the true gradient. This suggests that when the batch size falls within a practical optimal range, the batch size can give a good estimation of true gradient and training steps can serve as a reliable metric for tracking loss curves, allowing us to focus on other influential factors. However, to save resources and ensure stable training, batch sizes should not be excessively large (Figure~\ref{fig:loss_tokens_bsz} \emph{(Right)} when batch size is 2048).

Furthermore, our work addresses the second question: how do factors such as maximum learning rate, total training steps, and model size influence the annealing strategy? The optimal values of these factors are interdependent, with each influencing the selection of the others. For instance, previous studies \citep{Goyal2017AccurateLM, Hoffer2017TrainLG, You2019LargeBO, Granziol2020LearningRA} have primarily focused on the selection and scaling of maximum learning rates with respect to batch sizes, introducing strategies such as square-root scaling and linear scaling for the SGD optimizer. \citet{Li2024SurgePI} observed that for the Adam optimizer, the optimal learning rate initially increases and then decreases as the batch size grows. In this work, we go beyond individual factor selection and explore how these factors like the maximum learning rate influence the effectiveness of annealing strategies. We further propose a practical model-agnostic annealing strategy that adapts to different factors like maximum learning rate and model sizes.

Finally, we address the third question: what is the impact of the annealing strategy and LR scheduler on generalization? The key factors are the accumulated \textit{forward effect} of learning rates and the \textit{annealing momentum}, which is a measure of the kinetic effect of learning rate decay. Annealing momentum captures both the rate and magnitude at which the LR decreases during annealing, thus characterizing the convergence dynamics in the decay phase. The annealing ratio (\( R \)) is defined as the fraction of total training steps spent in the decay phase of the learning rate schedule: \(  R = \frac{T_{\text{decay}}}{T_{\text{total}}} \).  \citet{tissue2024scaling} proposed a scaling law to fit the loss curve across different schedulers throughout the training process, considering both forward effect and annealing degree. However, this formula assumes fixed batch sizes, and their forward term is not robust to variations in batch size. Moreover, their annealing momentum term relies on multiplicative accumulation, which can cause instability during calculation. In contrast, our approach uses integrals to compute both the forward and annealing momentum terms, which are more robust to batch size variations and irregular step counts. Additionally, we replace the multiplicative accumulation in the momentum term with an Adam-style momentum, mitigating the risk of instability and improving robustness to unstable training dynamics and hyperparameter tuning. We tested on WSD scheduler to investigate the relationship between optimal annealing ratio and factors such as total steps, then verified its transfer properties across different datasets.

Here are our key contributions:

1. \textit{Development of a Model-Agnostic Predictive Model}: We developed a robust predictive model for training dynamics, considering forward and annealing effects along with model size. Validated on both Dense and MoE models, it demonstrates an agnostic relationship and can accurately predict the impact of learning rate schedules.

2. \textit{Analysis of Key Factors Influencing Annealing Dynamics}: We have investigated how factors such as batch size, maximum learning rate, model architecture, and model size influence the annealing dynamics. We use training steps as a more reliable tracker for loss curves and explore the effects of factors like maximum learning rates and model characteristics on the annealing strategy.

3. \textit{Transferability of Annealing Strategy}: We find that the optimal annealing ratio can be transferred via specific principles. This suggests the existence of a consistent and predictable principle governing the annealing dynamics across various settings.

\section{Related Work}

\paragraph{Large Language Models}
Recently, Large Language Models (LLMs) such as GPT-4 \citep{achiam2023gpt}, LLaMA \citep{touvron2023llama}, Mistral \citep{jiang2023mistral}, DeepSeek \cite{bi2024deepseek, guo2025deepseek}, Longcat \cite{team2025longcat,chen2025mathematical,chen2025can} and Qwen \citep{yang2025qwen3technicalreport} have been significantly developed. Their performance in natural language understanding have gained great attention, with the number of training tokens, model sizes, and compute budgets scaling up to huge levels. However, there remain numerous aspects of their training dynamics and intermediate training loss curve that require further exploration, such as the impact of different annealing strategies on generalization scenarios.

\paragraph{Scaling Laws for LLMs}
Scaling laws \citep{Kaplan2020ScalingLF, Hoffmann2022TrainingCL} have been extensively studied, which revealed the power-law relationship between pre-training cross-entropy loss and factors such as the amount of training tokens, model size, and compute budget. Additionally, optimal hyperparameters \citep{mccandlish2018empirical, Hu2024MiniCPMUT, Li2024SurgePI} have been shown to scale with training loss according to specific patterns. Previous studies \citep{Yang2022TensorPV, Yang2023TensorPV,chen2025revisiting} indicate that certain hyperparameters, like batch size and sequence length, can be scaled across model depth or width. However, there is a lack of investigation into the whole training dynamics for a systematic analysis.

\paragraph{Learning Rate Schedulers for LLMs}
\citet{rae2021scaling, Hoffmann2022TrainingCL} recommend the cosine scheduler \citep{loshchilov2017sgdrstochasticgradientdescent} for LLM pre-training. However, it is unsuitable for continued training because its period depends on the training length. \citet{Zhai2021ScalingVT, Raffel2019ExploringTL, Hu2024MiniCPMUT} proposed WSD scheduler featuring a flexible warmup-stable-decay LR scheduling. Previous works \citep{tissue2024scaling, hagele2024scaling} have shown that its performance is comparable to the cosine scheduler and that scaling laws can also describe its final loss. \citet{Wen2024UnderstandingWL} introduced a variant of the WSD scheduler for continual learning to avoid performance drops after rewarming. \citet{defazio2024optimallineardecaylearning, schaipp2025surprisingagreementconvexoptimization} shows that the loss curves in large model training exhibit behavior similar to the performance bound derived from non-smooth convex optimization theory. In this work, we analyze the factors influencing the annealing strategy and explore its generalization across different configurations.

\section{Forward-Momentum Scaling Law}

Previous works \cite{Kaplan2020ScalingLF, Hoffmann2022TrainingCL} have proposed empirical scaling laws to predict the final loss for large language models:

\begin{equation}
    L = \frac{\lambda_N}{N^{\alpha_N}} + \frac{\lambda_D}{D^{\alpha_D}} + \sigma
\label{eq:loss_N_D_function}
\end{equation}

\noindent where \( N \) is the model size, and \( D \) is the number of training tokens. The parameters \( \alpha_N \), \( \alpha_D \), \( \lambda_N \), and \( \lambda_D \) are coefficients, and \( \sigma \) is the irreducible loss. However, this scaling law cannot capture differences in training loss curves when the same model is trained with the same number of tokens but different batch sizes or learning rate schedulers.

To better capture the discrepancies between different training settings, we tested training curves across various configurations with identical token counts and model sizes (Figure~\ref{fig:loss_tokens_bsz}). The training curves vary significantly when different batch sizes are used, even with the same maximum learning rate and scheduler. Yet we found that when tracking the loss curve using training steps, the curves tend to converge and become more alike.

This observation indicates that the trajectory of model optimization is more stable with respect to steps than tokens, especially in a practical batch size range, where training efficiency does not vary drastically. This aligns with the intuition that, under similar optimization dynamics (like the same scheduler), the number of parameter updates is a more direct indicator of training progress than the total number of tokens processed. The steps-based normalization implicitly captures the optimization view of training, treating each gradient update as a unit of progress, in contrast to token-based views which are more aligned with data consumption rather than model convergence.

Previous works \cite{Kaplan2020ScalingLF, Fedus2021SwitchTS} have shown that tracking loss with respect to training steps is practical under the assumptions of infinite data and stable, Adam-optimized training. Their experiments demonstrate that loss follows a power-law relationship with training steps, providing a theoretical foundation for our choice of step-based modeling. Therefore, we adopt training steps as the primary unit to analyze and compare training dynamics, which contributes a forward effect on loss decreasing. Based on Equation~\ref{eq:loss_N_D_function}, all training steps after the warm-up period are included:

\begin{equation}
    L(N, T) = \frac{\lambda_N}{N^{\alpha_N}} + \frac{\lambda_{T}}{{T}^{\alpha_T}} + \sigma
\label{eq:L_T_N_function}
\end{equation}

\noindent where $T$ is the training steps required to reach a targeted loss with fixed a batch size. And $\lambda_N, \alpha_{N}, \lambda_T, \alpha_T$ are all coefficients, $\sigma$ is the irreducible loss.

Besides the forward effects brought by training steps, the annealing phase (cooldown phase or decay phase) also plays a crucial role in loss reduction \cite{hagele2024scaling, Granziol2020LearningRA}. Learning rate annealing minimizes the oscillation of the parameter updates and brings the model closer to the minimum point. This behavior cannot be captured solely by the power-law term of training steps. Yet previous work \cite{tissue2024scaling} proposed an annealing term to describe the behavior during the annealing stage. Inspired by the empirical observation and previous works, we decided to incorporate a momentum term $M$ to describe the annealing behavior. The new formula is:

\begin{equation}
    L = \lambda_S \cdot S^{-\alpha} + \lambda_M \cdot M + L_0
\label{eq:Loss_S_M_function}
\end{equation}

\noindent where $L_0$, $\lambda_S$, $\lambda_M$ are coefficients. \(S\) is the integral of learning rate \( \eta \) with respect to steps (Equation~\ref{eq:S_term_definition}) and \(M\) is the integral of momentum with respect to steps (Equation~\ref{eq:M_term_definition}).

\begin{equation}
    S = \int_0^{T} \eta(t) \, dt
\label{eq:S_term_definition}
\end{equation}

\noindent where \( \eta(t) \) represents the learning rate as a function of the training step \( t \), and \( T \) is the total number of training steps.

\begin{figure*}[t]
\centering
\includegraphics[width=0.8\textwidth]{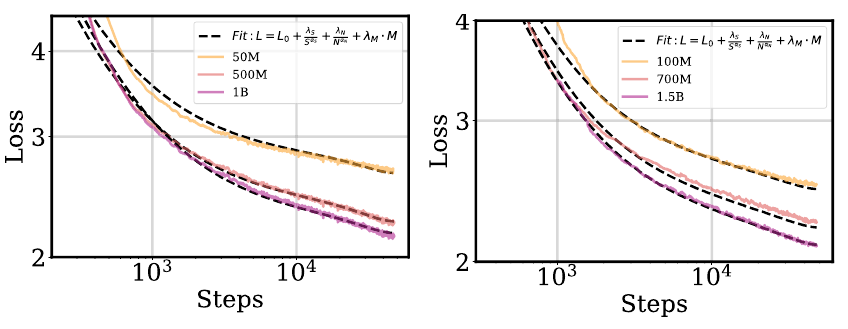}
\caption{Fitting results across model sizes. \emph{(Left)} Dense models (50M, 500M, 1B) with batch size 256 and learning rate 5e-5. \emph{(Right)} MoE models (100M, 700M, 1.5B) with batch size 256 and learning rate 2e-4. All use cosine learning rate scheduler.}
\label{fig:fitting_for_dense_moe_across_model_size}
\end{figure*}

\begin{figure*}[t]
\centering
\includegraphics[width=0.8\textwidth]
{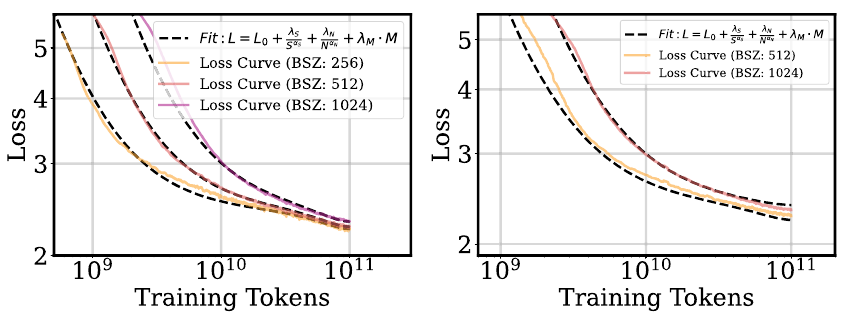}
\caption{Fitting results across different batch sizes using cosine scheduler. \emph{(Left)} Dense 500M models with batch sizes 256, 512, 1024 and max LR of 6e-4. \emph{(Right)} MoE 700M models with batch sizes 512 and 1024 using max LR of 2e-4.}
\label{fig:training_loss_tokens_bsz_lr}
\end{figure*}

\begin{equation}
    M = \int_0^{T} \text{momentum}(t) \, dt
\label{eq:M_term_definition}
\end{equation}
   
\noindent where \( \text{momentum}(t) \) represents the momentum as a function of training step \( t \), and \( T \) is the total number of training steps. Based on this definition, we can calculate $M$ using the following steps. Firstly, the momentum \( m_t \) and second moment \( v_t \) are updated at each step using the formulas:

\begin{equation}
    m_t = \beta_1 \cdot m_{t-1} + (1 - \beta_1) \cdot \Delta \eta_t  
\label{eq:m_t_definition}
\end{equation}

\begin{equation}
    v_t = \beta_2 \cdot v_{t-1} + (1 - \beta_2) \cdot (\Delta \eta_t)^2
\label{eq:v_t_definition}
\end{equation}

Then bias correction is applied to both moments:

\begin{equation}
    \hat{m}_t = \frac{m_t}{1 - \beta_1^t}, \quad \hat{v}_t = \frac{v_t}{1 - \beta_2^t}
\label{eq:m_t_v_t_correction}
\end{equation}

The cumulative momentum \( M_t \) is updated as:

\begin{equation}
    M_t = M_{t-1} + \frac{\hat{m}_t}{\sqrt{\hat{v}_t + \epsilon}}
\label{eq:cumulative_momentum}
\end{equation}

This method adjusts the influence of the first and second moments over time, providing a refined estimate of the momentum, and stabilizing updates with small \( \epsilon \) values. Thus, \(S\) captures the forward force to reduce the training loss, and \(M\) captures the loss reduction brought by the learning rate scheduler and annealing strategy.

Finally, we incorporate the model size term:

\begin{equation}
    L = \lambda_S \cdot S^{-\alpha_S} + \lambda_N \cdot N^{-\alpha_N} + \lambda_M \cdot M + L_0
\label{eq:Loss_S_N_M_E_function}
\end{equation}

\noindent 
where \(L\) is the training loss, \(M\) and \(S\) are the integrals of momentum and learning rate with respect to steps. \(L_0\), \(\lambda_S\), \(\lambda_M\), \(\lambda_N\), \(\alpha_S\), and \(\alpha_N\) are coefficients.

\section{Training Loss Curve Fitting}

Loss curves vary significantly depending on various hyperparameters. In this section, we investigate whether the loss curve can be accurately modeled across different training configurations within acceptable error margins, such as model sizes, batch sizes, and learning rate schedulers. We compare our Adam-style momentum formulation with multiplicative accumulations (CMMT) \citep{tissue2024scaling}, indicating Adam-style updates achieve the best stability and generalizability in transfer scenarios (more details shown in supplementary materials).

\subsection{Across Model Sizes Fitting}\label{subsec:model_size_fitting}
Equation~\ref{eq:loss_N_D_function} can be used to model final loss across different model sizes. However, to capture the full training dynamics, we propose a more comprehensive formulation: Equation~\ref{eq:Loss_S_N_M_E_function}, and we apply it to both MoE and Dense models.

As shown in Figure~\ref{fig:fitting_for_dense_moe_across_model_size}, we first control for batch size, total training steps, maximum learning rate, and scheduler type to ensure consistency. For Dense models (50M, 500M, and 1B), we use a fixed batch size of 256 and a cosine scheduler with a maximum learning rate of 5e-5. Our fitted loss curves achieve an average MAPE (Mean Absolute Percentage Error) below 2\%. Similarly, for MoE models (100M, 700M, and 1.5B), using the same batch size and a maximum learning rate of 2e-4, the MAPE loss remains consistently below 2\%. The sequence length is 8192. These results confirm that Equation~\ref{eq:Loss_S_N_M_E_function} effectively captures loss trends across model sizes, reinforcing its generalizability.

\subsubsection*{Observation 1: Model Size Dependence}

The loss curves exhibit a power-law dependence on model size, as described in 
Equation~\ref{eq:Loss_S_N_M_E_function}, enabling consistent extrapolation across different model scales.

\begin{figure*}[t]
\centering
\includegraphics[width=0.8\textwidth]{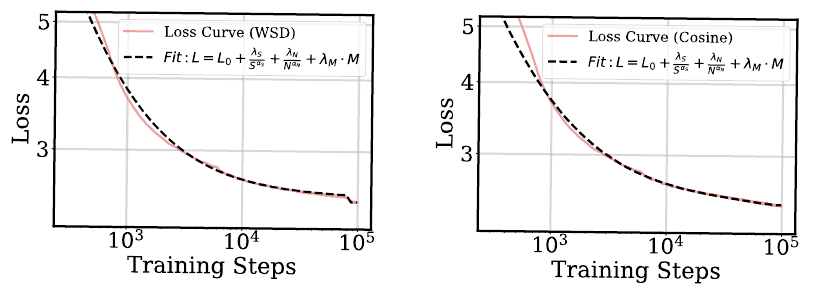}
\caption{Fitting results for 100M Dense models using different learning rate schedulers (Cosine and WSD) with the same total training steps and max LR. \emph{(Left)} Cosine scheduler loss curves used to predict those of WSD. \emph{(Right)} Vice versa.}
\label{fig:training_loss_tokens_cross_scheduler}
\end{figure*}

\subsection{Across Batch Sizes Fitting}\label{subsec:batch_size_fitting}

Training the same model with different batch sizes results in different training curves when using token-based loss tracking. However, our experiments show that when the batch size exceeds a certain threshold, the step-based training loss curves become close to each other.

The threshold corresponds to the optimal batch size (\(B_{opt}\)). \(B_{opt}\) is the minimum size that can approximate the true gradient, offering a good approximation while minimizing computational cost \cite{mccandlish2018empirical, Li2024SurgePI}. We assume that a wide range of batch sizes (greater than \(B_{opt}\)) with the same learning rate schedule produce similar loss v.s.steps curves. To verify our assumption, we fit the training curves using the same model size, maximum learning rate, and schedulers, varying only the batch size values (greater than \(B_{opt}\)). To ensure its generalization, we test this on both MoE and Dense models.

Firstly, we estimate the scaling laws between loss values and the optimal batch sizes. By examining the training curves across different batch sizes, we select and estimate the batch sizes that allow the model to use the minimum number of training tokens to reach a specific loss value \cite{Hu2024MiniCPMUT}.  We estimate that the optimal batch size for 500M Dense Model is about 293 (with sequence length as 8192), while that for 700M MoE is about 366 (with sequence length as 8192). After determining the corresponding batch size interval, we fit the selected training curves. The results show that step-based loss tracking is more effective for both (More details are shown in Figure~\ref{fig:training_loss_tokens_bsz_lr}).

\subsubsection*{Observation 2: Batch Size Dependence}

For a fixed model, a wide range of batch sizes larger than \(B_{\text{opt}}\), when paired with the same learning-rate schedule, produce nearly identical training curves. 
The optimal batch size \(B_{\text{opt}}\) follows a power-law relationship with the training loss:
\begin{equation}
	B_{\text{opt}} = \frac{\lambda_B}{L^{\alpha_B}}
	\label{eq:B_opt_Loss_function}
\end{equation}

\noindent where \(L\) is the training loss and \(B_{\text{opt}}\) is the optimal batch size. 
The parameters \(\lambda_B\) and \(\alpha_B\) are coefficients.

For MoE Models, \( \alpha_B \approx 3.430 \times 10^0 \) and \( \lambda_B \approx 4.390 \times 10^7 \), while for Dense Models, \( \alpha_B \approx 2.710 \times 10^0 \) and \( \lambda_B \approx 2.711 \times 10^7 \).

\subsection{Across Schedulers Fitting}\label{subsec:across_schedulers_fitting}

Based on previous work \citep{tissue2024scaling}, Equation~\ref{eq:Loss_S_M_function} focuses on capturing the impact of the learning rate schedule and annealing strategy. The power-law dependencies in \(S\) and \(N\) are consistent across different schedulers, while the momentum term \(M\) captures the specific behavior for each. It is optimistically expected that Equation~\ref{eq:Loss_S_M_function} can be transferable across different schedulers.

In our work, we primarily focus on two types of schedulers: the cosine scheduler and the WSD scheduler. For the cosine scheduler, the total number of steps is equal to the period of the scheduler. In contrast, the WSD scheduler consists of a warmup, constant, and linear decay phase. Its training loss curve typically ends with a sharp drop during the annealing stage. The formula for the WSD scheduler is:

\begin{equation}
    \eta(t) =
    \begin{cases}
    \eta_{\text{max}} \cdot \frac{t}{T_{\text{warmup}}}, & 0 \leq t < T_{\text{warmup}} \\
    \eta_{\text{max}}, & T_{\text{warmup}} \leq t < T_{\text{constant}} \\
    \eta_{\text{max}} \cdot \left(1 - \frac{t -T_{\text{constant}}}{T_{\text{decay}}}\right)^{\beta}, & T_{\text{constant}} \leq t < T_{\text{total}} \\
    \end{cases}
\label{eq:WSD_definition}
\end{equation}

\noindent where \(\beta = 1\) (Details shown in supplementary materials).

We verified that the fitting results for the cosine scheduler’s loss curve can predict those of the WSD scheduler, and vice versa ( Figure~\ref{fig:training_loss_tokens_cross_scheduler} and Table~\ref{tab:fitting_across_schedulers}). More detailed comparison is shown in supplementary materials.

\subsubsection*{Observation 3: Scheduler Dependence}

Different learning-rate schedulers influence the loss curve in a predictable manner, 
as captured by Equation~\ref{eq:Loss_S_M_function}.

\begin{table}[t]
\centering
\begin{tabular}{lcc}
\toprule
\textbf{Model Size} & \textbf{Cos Scheduler} & \textbf{WSD Scheduler} \\
\midrule
50M  & $0.456 \pm 0.061$ & $0.720 \pm 0.132$ \\
100M & $0.232 \pm 0.020$ & $0.410 \pm 0.019$ \\
500M & $0.453 \pm 0.038$ & $0.798 \pm 0.014$ \\
\bottomrule
\end{tabular}
\caption{MAPE (\%) loss prediction. "Cos Scheduler" refers to predicting the cosine scheduler's loss curve using the WSD-fitted model, and vice versa. Results are reported as mean~$\pm$~std over 3 random seeds.}
\label{tab:fitting_across_schedulers}
\end{table}

\begin{figure*}[t]
\centering
\includegraphics[width=0.9\textwidth]{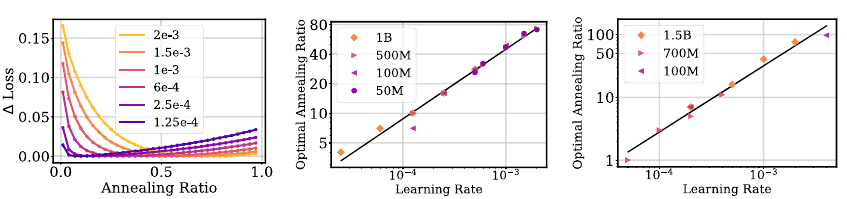}
\caption{Optimal annealing ratio across schedulers and maximum learning rates. \emph{(Left)} $\Delta_{Loss}$ vs. annealing ratio under different maximum learning rates for the 500M Dense model. \emph{(Middle)} Optimal annealing ratio vs. maximum learning rate across model sizes for Dense models. \emph{(Right)} Same plot for MoE models. All experiments use batch size 512, sequence length 8192, and 24k steps. \(\Delta_{Loss}\) is defined as gap between the final loss at each annealing ratio and the lowest final loss observed at \(R_{opt}\).}
    \label{fig:optimal_annealing_ratio_vs_lr}
\end{figure*}

\begin{figure*}[t]
\centering
\includegraphics[width=0.55\textwidth]{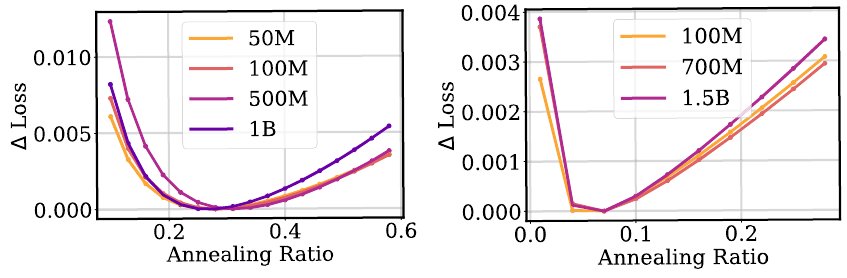}
\caption{
Transferability of optimal annealing ratio across model sizes.
\emph{(Left)} Dense models with batch size 512, sequence length 8192, 24k steps, and max LR of 4e-5.
\emph{(Right)} MoE models with the same batch size and steps, and max LR of 2e-4.
}
\label{fig:optimal_annealing_ratio_transfer_across_model_size}
\end{figure*}

\section{Transfer of Optimal Annealing Ratio}
It is crucial for schedulers to balance the forward effect of large learning rate with the convergence effect of annealing. A larger initial learning rate helps prevent the model from getting stuck in local minima. As training progresses and the model begins to converge, annealing stage allows model to refine its understanding of the data and capture more subtle, complex patterns. For the cosine scheduler, the optimal annealing strategy is to set the period of the scheduler equal to the total number of steps. However, for the WSD scheduler, selecting the optimal annealing ratio is more complex, which should consider the total steps, maximum learning rate, model sizes, model architectures and so on.

However, the annealing behavior across different configurations has not always been well understood yet. In this section, we will validate whether the optimal annealing ratio (\(R_{opt}\)) can be transferable across different training settings (maximum learning rates, model sizes, datasets, and training steps) to enable efficient and robust training.

\subsection{Across \textbf{\(LR_{max}\)} Transferability}

The maximum learning rate after warmup phase is crucial for training language models. \citet{Kaplan2020ScalingLF} suggests that smaller models can tolerate larger learning rates, while larger models require smaller learning rates to avoid divergence. And its impact on training becomes more significant when model size becomes larger. In our work, we fit the training curves for different maximum learning rates across each model size (as shown in Figure~\ref{fig:optimal_annealing_ratio_vs_lr} \emph{(Left)}). We then estimate the optimal annealing ratio for each corresponding maximum learning rate (Equation~\ref{eq:R_eta_equation}). 

\subsubsection*{Observation 1: $R_{\text{opt}}$ Scaling Across $\eta_{\max}$}

Empirically, the optimal annealing ratio follows a power-law relationship with the maximum learning rate:
\begin{equation}
	R_{\text{opt}} = \lambda_{\eta} \cdot \eta_{\max}^{\alpha_{\eta}}
	\label{eq:R_eta_equation}
\end{equation}

\noindent where $R_{\text{opt}}$ is the optimal annealing ratio and $\eta_{\max}$ is the maximum learning rate. $\lambda_{\eta}$ and $\alpha_{\eta}$ are coefficients.

This power-law behavior is consistent across different model sizes, regardless of whether the models are Dense or MoE (as shown in Figure~\ref{fig:optimal_annealing_ratio_vs_lr} \emph{(Middle)} and \emph{(Right)}). For Dense Models, \( \lambda_{\eta} \approx 5.996 \times 10^3 \) and \( \alpha_{\eta} \approx 7.09 \times 10^{-1} \). While for MoE Models, \( \lambda_{\eta} \approx 4.675 \times 10^4 \) and \( \alpha_{\eta} \approx 1.056 \).

\subsection{Across Model Transferability}

\begin{figure*}[t]
\centering
\includegraphics[width=1\textwidth]{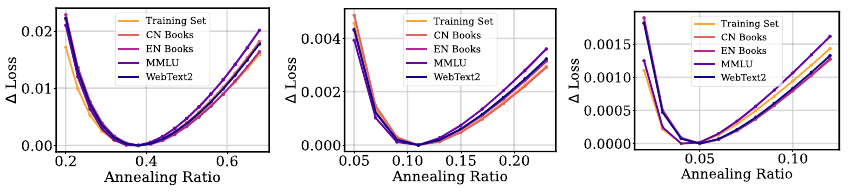}
\caption{
Optimal annealing ratio transferability for the 1B MoE model across different validation sets, under a fixed max learning rate of 2e-4 and batch size of 640. Each subplot shows $\Delta_{\text{Loss}}$ vs.\ annealing ratio at different total training steps. \emph{(Left)} 28.9k steps, \emph{(Middle)} 111k steps, \emph{(Right)} 227k steps.
}
\label{fig:optimal_annealing_ratio_transfer}
\end{figure*}

\citet{Yang2022TensorPV, Yang2023TensorPV} propose that the optimal hyperparameters for smaller models can be scaled and transferred to their larger counterparts. Thus, smaller models can serve as proxies for larger ones. Therefore, we assume that the optimal annealing ratios \(R_{opt}\) between small and large models can be transferred according to a certain principle.
 
We fit the coefficients across different model sizes with fixed maximum learning rate, total steps and batch size. Then the forward term and momentum term values are calculated to estimate the final losses at the end of training. We plot the gap (\(\Delta_{Loss}\)) between the final loss at each annealing ratio and the lowest final loss observed at \(R_{opt}\), to quantify the sensitivity of convergence to annealing choices across model sizes. As shown in Figure~\ref{fig:optimal_annealing_ratio_transfer_across_model_size}, the results show that \(R_{opt}\) can be transferred across different model sizes for both Dense and MoE Models. This strengthens the assumption that certain aspects of training dynamics are stable across architectures and sizes when hyperparameters are matched.

\subsubsection*{Observation 2: $R_{\text{opt}}$ Scaling Across Model Sizes}

Empirically, the optimal annealing ratio remains stable and transferable across different model sizes for Dense and MoE.

Noticeably, the performance gaps \(\Delta_{Loss}\) between different annealing ratios become larger when increasing the model size. Yet the optimal annealing ratio across different model sizes converges at the same value.

\subsection{Across Dataset Transferability}
\label{sec:across_dataset_transfer}

We further verify that the optimal annealing ratio (\(R_{opt}\)) predicted from training loss also generalizes across validation datasets with minimal degradation, suggesting that training dynamics capture generalization trends to a significant degree. We demonstrate the details and our findings using 1B MoE Models with WSD scheduler when maximum learning rate is 2e-4 and batch size is 640 with different steps. The absolute final loss error is within 0.003.

Figure~\ref{fig:optimal_annealing_ratio_transfer} shows that \(R_{opt}\) remains consistent across training and validation sets. Yet it is shown that the loss gaps (\(\Delta_{Loss}\)) can be more significant when we transfer the annealing ratio to validation sets. Besides, it is indicated that with the increasing of the total training steps, the optimal annealing ratio is decreasing. If we enter the annealing stage too much early, we risk overfitting. Moreover, the performance discrepancies (\(\Delta_{Loss}\)) between different ratios becomes smaller when increasing training steps. That means that longer training steps generally allow the model to explore the parameter space more thoroughly, leading to better generalization. It indicates a convergence when training a model for a large number of steps. 

\subsubsection*{Observation 3: $R_{opt}$ Scaling Across Datasets}

Empirically, the optimal annealing ratio remains consistent across training and validation sets.

\subsection{Across Steps Transferability}

\begin{figure}[h!]
\centering
\includegraphics[width=0.6\columnwidth]{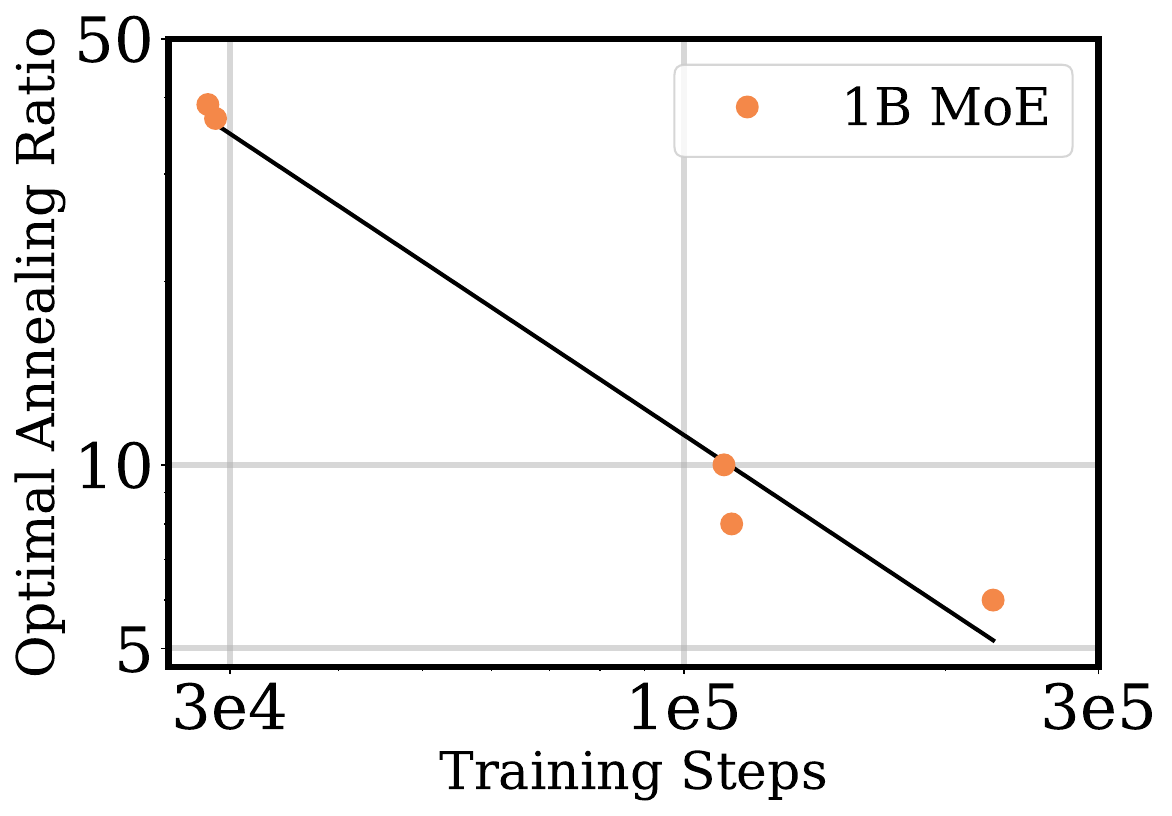}
\caption{
Optimal annealing ratio vs.\ training steps for 1B MoE models. 
Total steps are {30k, 100k, 300k}, all with fixed batch size 640 and maximum learning rate 2e-4.
}
\label{fig:optimal_annealing_ratio_vs_steps_1B_MoE}
\end{figure}

From previous section, we observed that the optimal annealing ratio ($R_{opt}$) decreases as the number of total training steps ($T$) increases. To quantify this relationship , we computed the optimal annealing ratio across different training steps and visualized the results (Equation~\ref{eq:R_T_equation}) in Figure~\ref{fig:optimal_annealing_ratio_vs_steps_1B_MoE}.

\subsubsection*{Observation 4: $R_{\text{opt}}$ Scaling Across Steps $T$}

Empirically, $R_{\text{opt}}$ and $T$ follows a power-law form:
\begin{equation}
	R_{\text{opt}} = \lambda_T \cdot T^{\alpha_T}
	\label{eq:R_T_equation}
\end{equation}

\noindent where $R_{\text{opt}}$ represents the optimal annealing ratio, and $T$ denotes the number of training steps. 
The parameters $\lambda_T$ and $\alpha_T$ are data-dependent coefficients.

Specifically, for 1B MoE models, we estimate \( \lambda_T \approx 5.987 \times 10^5 \) and \( \alpha_T \approx -9.46 \times 10^{-1} \). This implies that when using the WSD scheduler with a batch size within the optimal range, the annealing strategy can be directly inferred from the total training steps or maximum learning rate.

\section{Conclusion}
In this paper, we demonstrate that the training steps could be a better tracker for loss curves than total training tokens, especially when the batch size is larger than the threshold (the optimal batch size). Then we investigate three key factors influencing training loss curves: the cumulative forward effect, the annealing momentum over training steps and model size. Then we validate the prediction accuracy across various scenarios. Finally, through estimating the accumulated momentum value, we examine its transferability across varying total training steps, model sizes, maximum learning rates, and validation datasets. These results show that annealing policies are transferable and predictable across configurations, enabling efficient LLM training.

\section*{Acknowledgments}
This work was partially supported by research grants from the Research Grants Council (RGC) of the Hong Kong Special Administrative Region, China (Project Nos. R6005-24 and AoE/E-601/24-N), and the Hong Kong Joint Research Scheme (JRS) of the National Natural Science Foundation of China (NSFC)/RGC (Project No. N\_HKUST654/24). We would like to thank Ziqing Xu for his insightful advice on the implementation details. This work was also supported by the Hong Kong University of Science and Technology (HKUST) CSE PhD Fellowship.

\bibliography{aaai2026}

\appendix

\section{Experimental Settings}
\subsection{Training Details}
\label{subsect:Experimental_Settings}
The details about the number of layers, attention heads, hidden dimensions, and other relevant information for Dense Models and MoE Models are listed in Table~\ref{tab:dense_arch} and Table~\ref{tab:moe_arch}.

During cosine scheduler \citep{loshchilov2017sgdrstochasticgradientdescent} training, AdamW adopts parameters \(\beta_1 = 0.9\) and \(\beta_2 = 0.95\). Following the previous work \cite{Hoffmann2022TrainingCL}, the maximum learning rate for the cosine scheduler is \(1.5 \times 10^{-3}\) for smaller models and \(2 \times 10^{-4}\) for larger ones. The cosine scheduler with a 10x learning rate decay was implemented throughout the training process. We adopt smoothing with a 50-step window length to smooth the training curve. The WSD scheduler used in 1B MoE Model experiments has a maximum learning rate of \(2 \times 10^{-4}\) and 0.1 annealing ratios. The total training steps vary from 29k to 227k. The sequence length used in experiments is 8192.

Specifically, the training dataset for Dense and MoE Models is The Pile \cite{gao2020pile}. The validation datasets include MMLU \cite{Hendrycks2020MeasuringMM}, OpenWebtext2 \cite{Gokaslan2019OpenWeb}, CN Books, and EN Books \cite{Zhu2015AligningBA}.

\begin{table}[h]
	\centering
	\setlength{\tabcolsep}{3pt}
	\renewcommand{\arraystretch}{1.2}
	\begin{tabular}{cccccc}
		\toprule
		\textbf{Model Size} & \textbf{Params} & \textbf{Hid. Size} & \textbf{Layers} & \textbf{Heads} & \textbf{FFN} \\
		\midrule
		50M  & 47.7M   & 512  & 14 & 4  & 1536  \\
		100M & 113.3M  & 768  & 16 & 6  & 2048  \\
		500M & 487.6M  & 1280 & 24 & 10 & 3584  \\
		1B   & 936.3M  & 1664 & 28 & 13 & 4480  \\
		7B   & 7.03B   & 4096 & 32 & 32 & 11008 \\
		\bottomrule
	\end{tabular}
	\caption{Architecture details of the Dense Models used in our experiments.}
	\label{tab:dense_arch}
\end{table}
\begin{table}[h]
	\centering
	\setlength{\tabcolsep}{3pt}  
	\renewcommand{\arraystretch}{1.2}
	\begin{tabular}{cccccc}
		\toprule
		\textbf{Model Size} & \textbf{Act. Params} & \textbf{Hid.} & \textbf{Layers} & \textbf{Heads} & \textbf{FFN} \\
		\midrule
		200M  & 183M  & 640  & 12 & 8  & 1792 \\
		500M  & 485M  & 960  & 14 & 12 & 2688 \\
		700M  & 692M  & 1280 & 16 & 16 & 3456 \\
		1B    & 987M  & 1440 & 20 & 18 & 4032 \\
		1.5B  & 1.45B & 1600 & 24 & 20 & 4352 \\
		7B    & 6.8B  & 2240 & 30 & 28 & 6272 \\
		\bottomrule
	\end{tabular}
	\caption{Architecture details of the MoE models used in our experiments (all top-3 routing).}
	\label{tab:moe_arch}
\end{table}

\subsection{Fitting Procedure and Error Estimation}
\label{subsec:fitting_details}

To estimate the parameters of our scaling laws, we use robust nonlinear regression with the \texttt{L-BFGS-B} optimizer from \texttt{scipy.optimize.minimize}. The objective is to minimize the Huber loss with threshold \(\delta = 10^{-3}\) between predicted and observed losses, ensuring robustness to outliers in the presence of noisy training dynamics. The maximum iteration is 1000.

Each training loss curve is smoothed using a moving average with a 50-step window before fitting. All results are based on single-run trajectories unless stated otherwise. We report the Mean Absolute Percentage Error (MAPE) to evaluate the prediction accuracy. 

This procedure is used consistently in the fitting process across model sizes, batch sizes, or schedulers.

\section{Additional Theoretical Background}

\subsection{Linear Decay in Learning Rate Schedules}
\citet{zamani2023exactconvergencerateiterate, defazio2024optimallineardecaylearning, schaipp2025surprisingagreementconvexoptimization} demonstrate that linear decay schedules achieve the optimal convergence rate for the last iterate in stochastic convex optimization problems with Lipschitz continuity assumptions. This makes them particularly effective for non-stationary optimization problems, such as LLM training. While LLM training is not strictly convex, empirical evidence suggests that LLMs exhibit behavior that closely mirrors the convergence patterns observed in convex optimization problems, particularly regarding the effectiveness of linear decay schedules. \citet{schaipp2025surprisingagreementconvexoptimization} notably show that linear decay schedules match the convergence behavior predicted by convex optimization theory, making them a robust choice for training LLMs. These insights inspired the use of linear decay in our work.

\subsection{Last-Iterate Bounds of SGD}
\label{sec:theory-sgd}

In this section, we introduce theoretical results from previous work~\citep{defazio2024optimallineardecaylearning} which provide a foundation for understanding the benefits of linear decay schedules for stochastic gradient descent (SGD). Their analysis introduces a reduction from regret bounds to convergence of the last iterate, thereby enabling direct bounding of $f(x_T) - f(x^\star)$ without averaging.

\subsubsection{Additive Reduction to Last Iterate Loss}

Their main result (Theorem~\ref{thm:additive}) shows that if a sequence $z_1, \ldots, z_T$ achieves low regret with respect to gradient directions, then a linearly interpolated sequence $x_t$ constructed from it satisfies a last-iterate loss bound.

\begin{theorem}[Additive Reduction to Last Iterate]
	Let $z_1,\dots,z_T$ be an arbitrary sequence of vectors and $\omega_1,\dots,\omega_T$ be non-negative weights. Define $\Delta_t = z_{t+1} - z_t$ and $x_1 = z_1$, with update rule:
	\begin{align*}
		x_{t+1} = x_t + \frac{\omega_{t+1:T}}{\omega_{1:T}} \Delta_t,
	\end{align*}
	then for any comparator $\cmp$:
	\begin{align*}
		\mathbb{E}[f(x_T) - f(\cmp)] \leq \frac{1}{\omega_{1:T}} \sum_{t=1}^T \mathbb{E}[\omega_t \langle g_t, z_t - \cmp \rangle].
	\end{align*}
	\label{thm:additive}
\end{theorem}

\subsubsection{Implication for Linear Decay Schedule}

A direct corollary of this theorem arises when we set $\omega_t = 1$ and assume the standard regret bound $\sum_{t=1}^T \langle g_t, z_t - \cmp \rangle \le DG\sqrt{T}$. Then by choosing:
\begin{align}
	x_{t+1} = x_t - \eta_t g_t, \quad \eta_t = \frac{D}{G\sqrt{T}}\left(1 - \frac{t}{T}\right), \label{eq:linear-decay}
\end{align}
we obtain the last-iterate convergence bound:
\begin{align}
	\mathbb{E}[f(x_T) - f(x^\star)] \le \frac{DG}{\sqrt{T}}.
\end{align}

This result demonstrates that a simple linearly decaying learning rate achieves the same convergence rate as iterate averaging but returns the final point $x_T$ rather than a mean of past iterates.

\subsection{Convex Last-Iterate Convergence for Cooldown}

\citet{schaipp2025surprisingagreementconvexoptimization} provide a rigorous theoretical analysis of last-iterate convergence rates for scheduled SGD under convex objectives. Consider the update rule:
\begin{equation}
	x_{t+1} = x_t - \gamma \eta_t g_t, \quad g_t \in \partial f(x_t, s_t),
\end{equation}
with a base learning rate $\gamma$ and schedule $\eta_t$.

Under convexity and a uniform bound on gradient norms, they prove the following bound for the expected last-iterate error:
\begin{equation}
	\E[f(x_T) - f(x_\star)] \leq \frac{\mathcal{T}_1}{\gamma} + \gamma \cdot \mathcal{T}_2,
\end{equation}
where $\mathcal{T}_1$ and $\mathcal{T}_2$ are functions of the schedule $\{\eta_t\}$, the gradient norm bound $G$, and the distance $D := \|x_1 - x_\star\|$.

The optimal base learning rate is given analytically as:
\begin{equation}
	\gamma^\star = \sqrt{\frac{\mathcal{T}_1}{\mathcal{T}_2}},
\end{equation}
yielding the minimal last-iterate bound:
\begin{equation}
	\E[f(x_T) - f(x_\star)] \leq 2 \sqrt{\mathcal{T}_1 \mathcal{T}_2}.
\end{equation}

\begin{table*}[h]
	\centering
	\begin{tabular}{@{}ll@{}}
		\toprule
		\textbf{Equation} & \textbf{Interpretation} \\
		\midrule
		$L = \lambda_N N^{-\alpha_N} + \lambda_D D^{-\alpha_D} + \sigma$ & Original Scaling Law~\citep{Kaplan2020ScalingLF, Hoffmann2022TrainingCL} \\
		$L = \lambda_N N^{-\alpha_N} + \lambda_T T^{-\alpha_T} + \sigma$ & Step-based Scaling Law (fixed batch size)~\citep{Kaplan2020ScalingLF} \\
		$L = \lambda_S S^{-\alpha_S} + \lambda_N N^{-\alpha_N} + \lambda_M M + L_0$ & Forward-Momentum Scaling Law (this work) \\
		$B_{\text{opt}} = \lambda_B \cdot L^{-\alpha_B}$ & Scaling of Optimal Batch Size w.r.t. Training Loss \\
		$R_{\text{opt}} = \lambda_\eta \cdot \eta_{\max}^{\alpha_\eta}$ & Annealing Ratio Scaling with $\eta_{\max}$ (WSD Linear Decay) \\
		$R_{\text{opt}} = \lambda_T \cdot T^{\alpha_T}$ & Annealing Ratio Scaling with Training Steps (WSD Linear Decay) \\
		\bottomrule
	\end{tabular}
	\caption{Summary of key scaling laws and their interpretations.}
	\label{tab:scaling_summary}
\end{table*}

\subsubsection{Comparison Between Constant and WSD Schedules}

They further compare constant schedules ($\eta_t=1$) with the recently proposed Warmup-Stage Decay (WSD) schedule. 
For constant schedules, the suboptimality bound includes a logarithmic term $H_{T-1} := \sum_{k=1}^{T-1} \frac{1}{k} \approx \ln T$ (harmonic number), which grows slowly but unbounded with $T$. While for WSD, the logarithmic growth disappears, and the bound becomes:
\begin{equation}
	\E[f(x_T) - f(x_\star)] \lesssim \frac{DG}{\sqrt{T}} \cdot \sqrt{C(\beta)},
\end{equation}
where $C(\beta)$ is independent of $T$ and depends only on the cooldown ratio $\beta$.

This provides theoretical support for empirical findings that WSD improves performance in later decay stage without harming training stability.

\section{Comparative Analysis of Momentum Terms}
\label{subsec:fitting_accuracy}
\subsection{Preliminaries}
\label{sec:prelim}

\paragraph{Annealing ratio.}
We define the annealing ratio as the fraction of training spent in the decay phase of the learning-rate schedule:
\[
R \;=\; \frac{T_{\mathrm{decay}}}{T_{\mathrm{total}}}\in(0,1],\qquad
R_{\mathrm{opt}}=\arg\min_R L_{\mathrm{final}}(R).
\]

\subsection{Formal Definition}

In our work, we propose a scaling law (Equation~\ref{eq:Loss_S_N_M_E_function}) that characterizes the training loss \(L\) as a function of (i) the cumulative forward effect induced by training steps accumulation, (ii) the accumulated effect of annealing momentum during cooldown (or decay), and (iii) model size:

\begin{equation}
	L = \lambda_S \cdot S^{-\alpha_S} + \lambda_N \cdot N^{-\alpha_N} + \lambda_M \cdot M + L_0
	\label{eq:Loss_S_N_M_E_function}
\end{equation}

\noindent 
where \(L\) is the training loss, \(M\) and \(S\) are the integrals of momentum and learning rate with respect to steps. \(L_0\), \(\lambda_S\), \(\lambda_M\), \(\lambda_N\), \(\alpha_S\), and \(\alpha_N\) are coefficients.

\(S\) is the integral of learning rate \( \eta \) with respect to steps (Equation~\ref{eq:S_term_definition}) and \(M\) is the integral of momentum with respect to steps (Equation~\ref{eq:M_term_definition}).

\begin{equation}
	S = \int_0^{T} \eta(t) \, dt
	\label{eq:S_term_definition}
\end{equation}

\noindent where \( \eta(t) \) represents the learning rate as a function of the training step \( t \), and \( T \) is the total number of training steps.

To characterize the \emph{kinetic effect} of annealing, we use annealing momentum to measure both the \emph{rate} and \emph{magnitude} at which the LR decreases during annealing and serves as a proxy for accumulated curvature exposure.

\begin{equation}
	M = \int_0^{T} \text{momentum}(t) \, dt
	\label{eq:M_term_definition}
\end{equation}

\noindent where \( \text{momentum}(t) \) represents the momentum as a function of training step \( t \), and \( T \) is the total number of training steps. 

To compute \(M\), we follow an Adam-style momentum update. The momentum \( m_t \) and second moment \( v_t \) are updated at each step using the formulas:

\begin{equation}
	m_t = \beta_1 \cdot m_{t-1} + (1 - \beta_1) \cdot \Delta \eta_t  
	\label{eq:m_t_definition}
\end{equation}

\begin{equation}
	v_t = \beta_2 \cdot v_{t-1} + (1 - \beta_2) \cdot (\Delta \eta_t)^2
	\label{eq:v_t_definition}
\end{equation}

Then bias correction is applied to both moments:

\begin{equation}
	\hat{m}_t = \frac{m_t}{1 - \beta_1^t}, \quad \hat{v}_t = \frac{v_t}{1 - \beta_2^t}
	\label{eq:m_t_v_t_correction}
\end{equation}

The step-wise contribution to the cumulative momentum is then:

\begin{equation}
	M_t = M_{t-1} + \frac{\hat{m}_t}{\sqrt{\hat{v}_t + \epsilon}}
	\label{eq:cumulative_momentum}
\end{equation}

Finally, we define the Adam-Style Momentum Term (ASMT) as:
\begin{equation}
	M_A = M_{T}
	\label{eq:ASTM}
\end{equation}

\noindent where \(T\) is the final step.

Specifically, when the model size is held constant, Equation~\ref{eq:Loss_S_N_M_E_function} reduces to:

\begin{equation}
	L = \lambda_S \cdot S^{-\alpha} + \lambda_M \cdot M + L_0
	\label{eq:Loss_S_M_function}
\end{equation}

\noindent where $L_0$, $\lambda_S$, $\lambda_M$ are coefficients.

We compare the performance of our proposed Adam-Style Momentum Term (ASMT) with the Cumulative Multiplicative Momentum Term (CMMT) introduced in prior work \citep{tissue2024scaling}. The CMMT is defined as:

\begin{equation}
	M_{C} = \sum_{i=1}^{s} \sum_{k=1}^{i} (\eta_{k-1} - \eta_k) \cdot \lambda^{i-k}
\end{equation}

\noindent where \( \eta_i \) is the learning rate at step \( i \), and \( \lambda \) is a hyperparameter representing the decay factor for learning rate annealing momentum, typically ranging from 0.99 to 0.999. \( M_{C} \) represents the learning rate annealing area, capturing the cumulative effect of learning rate changes over time during training.

We summarize all scaling equations and formulas introduced or mentioned in this work in Table~\ref{tab:scaling_summary}, along with their interpretations.

In this section, we compare the performance of our proposed Adam-Style Momentum Term (ASMT) with the Cumulative Multiplicative Momentum Term (CMMT) introduced in prior work \citep{tissue2024scaling}, under momentum decay values of $\lambda = 0.999$ and $\lambda = 0.99$.

\subsection{Fitting process across model sizes}
During the fitting process across model sizes, we fit Equation~\ref{eq:Loss_S_N_M_E_function} across model sizes with fixed batch sizes, learning rate schedulers, and maximum learning rates for both Dense and MoE models.

For Dense models, the model sizes include 50M, 100M, 500M, and 1B, while for MoE models, the model sizes include 100M, 700M, and 1.5B. In Table~\ref{tab:avg-mape}, for Dense models, the batch sizes and maximum learning rates are shown (with a sequence length of 8192), and the scheduler is the Cosine Scheduler. For MoE models, the batch sizes and maximum learning rates are shown (with a sequence length of 8192), and the scheduler is also the Cosine Scheduler. This table evaluates robustness of ASMT and Equation~\ref{eq:Loss_S_N_M_E_function} to model size variation.

\begin{table}[ht]
	\centering
	\setlength{\tabcolsep}{1.4pt}
	\renewcommand{\arraystretch}{1.2}
	\begin{tabular}{lcccccc}
		\toprule
		\textbf{Model} & \textbf{$LR_{max}$} & \textbf{Batch Size} & ASMT  & \makecell{CMMT\\($\lambda=0.999$)} & \makecell{CMMT\\($\lambda=0.99$)} \\
		\midrule
		Dense & 1.25e-4 & 512 & 0.402 & 0.413 & 0.403\\
		Dense & 5e-5 & 512 & 0.579 & 0.943 & 0.887\\
		Dense & 5e-5 & 1024 & 0.565 & 0.566 & 0.567 \\
		MoE   & 2e-4 & 512 & 0.957 & 1.039 & 0.750\\
		MoE   & 2e-3 & 512 & 0.574 & 0.487& 0.525  \\
		MoE   & 1e-4 & 512 & 0.341 & 0.373 & 0.356 \\
		\bottomrule
	\end{tabular}
	\caption{Average Mean Absolute Percentage Error, MAPE (\%), when fitting across model sizes for Dense and MoE models.}
	\label{tab:avg-mape}
\end{table}

\subsection{Fitting process across batch sizes} During the fitting procedure across batch sizes, we fit Equation~\ref{eq:Loss_S_M_function} across batch sizes with fixed model sizes, learning rate schedulers, and maximum learning rates for both Dense and MoE models. 

\begin{table}[ht]
	\centering
	\setlength{\tabcolsep}{3pt}
	\renewcommand{\arraystretch}{1.2}
	\begin{tabular}{lccccc}
		\toprule
		\textbf{Model} & \textbf{Model Size} & \textbf{ASMT} & \makecell{CMMT\\($\lambda=0.999$)} & \makecell{CMMT\\($\lambda=0.99$)} \\
		\midrule
		\multirow{4}{*}{Dense} 
		& 50M   & 0.544 & 1.293 & 0.552\\
		& 100M  & 0.529 & 1.487 & 0.538\\
		& 500M  & 0.646 & 1.966 & 0.660\\
		& 1B    & 0.674 & 2.151 & 0.681\\
		\midrule
		\multirow{4}{*}{MoE} 
		& 100M  & 1.223 & 1.467 & 1.342\\
		& 700M  & 1.561 & 1.896 & 1.752\\
		& 1.5B  & 0.607 & 1.079 & 0.736\\
		\bottomrule
	\end{tabular}
	\caption{Mean Absolute Percentage Error, MAPE (\%), between predicted loss and true loss across batch sizes.}
	\label{tab:mape-results}
\end{table}

We compare the results between Adam-Style Momentum Term (ASMT) and Cumulative Multiplicative Momentum Term (CMMT) with $\lambda = 0.999$ and $\lambda = 0.99$. In Table~\ref{tab:mape-results}, for Dense models, the batch sizes include 256, 512, and 1024 (with sequence length 8192); the maximum learning rate is 5e-5, and the scheduler is the Cosine Scheduler. For MoE models, the batch sizes include 512 and 1024 (with sequence length 8192); the maximum learning rate is 2e-4, and the scheduler is also the Cosine Scheduler. This table evaluates robustness of ASMT and Equation~\ref{eq:Loss_S_M_function} to batch size variation.

\subsection{Fitting process across schedulers} 
\begin{table}[h!]
	\centering
	\setlength{\tabcolsep}{1.5pt}
	\renewcommand{\arraystretch}{1.3}
	\begin{tabular}{lccc}
		\toprule
		\textbf{Momentum} & \textbf{Model Size} & \textbf{Cos Scheduler} & \textbf{WSD Scheduler} \\
		\midrule
		\multirow{3}{*}{\shortstack{CMMT\\($\lambda=0.999$)}}
		& 50M   & 1.886 & 0.838 \\
		& 100M  & 2.251 & 0.422 \\
		& 500M  & 0.417 & 0.784 \\
		\midrule
		\multirow{3}{*}{\shortstack{CMMT\\($\lambda=0.99$)}}
		& 50M   & 0.539 & 0.777 \\
		& 100M  & 0.202 & 0.404 \\
		& 500M  & 0.379 & 0.657 \\
		\midrule
		\multirow{3}{*}{ASMT}
		& 50M   & 0.456 & 0.720 \\
		& 100M  & 0.232 & 0.410 \\
		& 500M  & 0.453 & 0.798 \\
		\bottomrule
	\end{tabular}
	\caption{MAPE (\%) for loss prediction. "Cos Scheduler" refers to predicting the cosine scheduler's loss curve using a model fitted on WSD scheduler data, and vice versa.}
	\label{tab:fitting_across_schedulers_cmmt}
\end{table}

During the fitting procedure across schedulers with fixed model size, we fit the loss curve of one scheduler using Equation~\ref{eq:Loss_S_M_function} and predict the loss curve of another.

In Table~\ref{tab:fitting_across_schedulers_cmmt}, we compare the performance of ASMT with the Cumulative Multiplicative Momentum Term (CMMT) under different $\lambda$ values (0.99 and 0.999). The results show that ASMT generally outperforms CMMT when $\lambda = 0.999$, while in some cases performing slightly worse than CMMT when $\lambda = 0.99$. Given that prior work does not offer clear guidance on how to fine-tune $\lambda$, ASMT demonstrates greater stability and robustness to hyperparameter selection. This result supports its transferability across schedulers.

Overall, we compared different momentum formulations, including multiplicative accumulations (CMMT) and Adam-Style Momentum Term (ASMT), and found that Adam-style updates achieve the best stability and generalizability in transfer scenarios.

\section{Scope and Limitation}
Our transferability claims hold \emph{within the same architecture} (Dense or MoE). We do not claim direct transfer \emph{across} Dense and MoE. Experiments span up to 7B parameters due to compute limits; extending to 10B+ and broader schedulers is left for future work.

\end{document}